# Statistical Decisions Using Likelihood Information Without Prior Probabilities

Phan H. Giang and Prakash P. Shenoy

University of Kansas School of Business
1300 Sunnyside Ave., Summerfield Hall, Lawrence, KS 66045-7585, USA
{phgiang,pshenoy} @ku.edu

## Abstract

This paper presents a decision-theoretic approach to statistical inference that satisfies the Likelihood Principle (LP) without using prior information. Unlike the Bayesian approach, which also satisfies LP, we do not assume knowledge of the prior distribution of the unknown parameter. With respect to information that can be obtained from an experiment, our solution is more efficient than Wald's minimax solution. However, with respect to information assumed to be known before the experiment, our solution demands less input than the Bayesian solution.

## 1 Introduction

The Likelihood Principle (LP) is one of the fundamental principles of statistical inference [8, 5, 3, 2]. A statistical inference problem can be formulated as follows. We are given a description of an experiment in the form of a partially specified probability model that consists of a random variable $Y$ which is assumed to follow one of the distributions in the family $\mathcal{F} = \{P_\theta | \theta \in \Omega\}$. The set of distributions is parameterized by $\theta$ whose space is $\Omega$. Suppose we observe $Y = y$, an outcome of the experiment. What can we conclude about the true value of the parameter $\theta$?. Roughly speaking, LP holds that all relevant information from the situation is encoded in the likelihood function on the parameter space. In practice, likelihood information is used according to the maximum likelihood procedure (also refered to as the maximum likelihood principle or method) whereby the likelihood assigned to a set of hypotheses is often taken to be the maximum of the likelihoods of individual hypothesis in the set. The power and importance of this procedure in statistics is testified by the following quote from Lehman [14]:

Although the maximum likelihood principle is not based on any clearly defined optimum consideration, it has been very successful in leading to satisfactory procedures in many specific problems. For wide classes of problem, maximum likelihood procedure have also been shown to possess various asymptotic optimum properties as the sample size tends to infinity.

A major approach to statistical inference is the decision-theoretic approach in which the statistical inference problem is viewed as a decision problem under uncertainty. This view implies that every action taken about the unknown parameter has consequences depending on the true value of the parameter. For example, in the context of an estimation problem, an action can be understood as the estimate of the parameter; or in the context of a hypothesis testing problem an action is a decision to *accept* or *reject* an hypothesis. The consequences of actions are valued by their utility or loss. A solution to a decision problem is selected according to certain theories. Two widely used decision theories are Wald's *minimax* and the expected utility maximization. The latter is appropriate for Bayesian statistical problems in which there is sufficient information to describe the posterior probability distribution on the hypothesis space. Bayesian approach agrees with LP. It holds that the relevant information from the experiment is indeed contained in the likelihood function. However, in addition to the experimental data, Bayesian statistics assumes the existence of a prior probability distribution, which summarizes the information about the unknown parameter *before* the experiment is conducted. This prior probability assumption is probably the most contentious topic in statistics.

In [10, 11], we presented an axiomatic approach to decision making where uncertainty about the true state of nature is expressed by a possibility function. Possibility theory is a relatively recent calculus for describ-ing uncertainty. Possibility theory has been developed and used mostly within the AI community. It was first proposed in late 1970s by Zadeh [23].

In this paper, we take the position of LP. In particular, we assume that likelihood information, as used in the maximum likelihood procedure, faithfully represents all relevant uncertainty about the unknown parameter. We show that such information is a possibility measure. Thus, our decision theory for possibility calculus is applicable to the problem of statistical inference. We argue that our approach satisfies LP. But, in contrast with the Bayesian approach, we do not assume any knowledge of the prior distribution of the unknown parameter. We claim that our solution is more information efficient than Wald's minimax solution, but demands less input than the Bayesian solution.

## 2 Likelihood Information and Possibility Function

Let us recall basic definitions in possibility theory. A possibility function is a mapping from the set of possible worlds $S$ to the unit interval $[0,1]$

$$\pi : S \to [0,1] \text{ such that } \max_{\omega \in S} \pi(\omega) = 1 \quad (1)$$

Possibility for a subset $A \subseteq S$ is defined as

$$\pi(A) = \max_{\omega \in A} \pi(\omega) \quad (2)$$

A conditional possibility[1] is defined for $A, B \subseteq S, \pi(A) \neq 0$

$$\pi(B|A) = \frac{\pi(A \cap B)}{\pi(A)} \quad (3)$$

Compared to probability functions, the major difference is that the possibility of a set is the maximum of possibilities of elements in the set.

Given a probabilistic model $\mathcal{F}$ and an observation $y$, we want to make inference about the unknown parameter $\theta$ which is in space $\Omega$.

The likelihood concept used in modern statistics was coined by R. A. Fisher who mentioned it as early as 1922 [9]. Fisher used likelihood to measure the "mental confidence" in competing scientific hypotheses as a result of a statistical experiment (see [8] for a detailed account). Likelihood has a puzzling nature. For a single value $\theta_0$, the likelihood is just $P_{\theta_0}(y)$, the probability of observing $y$ if $\theta_0$ is in fact the true value of the parameter. One can write $P_{\theta_0}(y)$ in the form of a conditional probability: $P(Y = y|\theta = \theta_0)$. The latter notation implies that there is a probability measure on parameter space $\Omega$. This is the case for the Bayesian approach. In this paper, we do not assume such a probability measure. So we will stick with the former notation. For each value $\theta \in \Omega$, there is a likelihood quantity. If we view the set of likelihood quantities as a function on the parameter space, we have a likelihood function. To emphasis the fact that a likelihood function is tied to data $y$ and has $\theta$ as the variable, the following notation is often used:

$$lik_y(\theta) = P_\theta(y) \quad (4)$$

Thus, a likelihood function is determined by a partially specified probabilistic model and an observation. It is important to note that the likelihood function can no longer can be interpreted as a probability function. For example, integration (or summation) of the likelihood function over the parameter space, in general, does not sum up to unity.

The Likelihood Principle (LP) states that all information about $\theta$ that can be obtained from an observation $y$ is contained in the likelihood function for $\theta$ given $y$, up to a proportional constant. LP has two powerful supporting arguments [3].

First, it is well known that likelihood function $lik_Y(\theta)$ is a minimal sufficient statistic for $\theta$. Thus, from the classical statistical point of view, the likelihood contains all information about $\theta$. Second, Birnbaum, in a seminal article published in 1962 [4], showed that the Likelihood Principle is logically equivalent to the combination of two fundamental principles in statistics: the principle of conditionality and the principle of sufficiency. The principle of conditionality says that only the actual observation is relevant to the analysis and not those that potentially could be observed. The principle of sufficiency says that in the context of a given probabilistic model, a sufficient statistic contains all relevant information about the parameter that can be obtained from data.

Let us define an "extended" likelihood function $Lik_y : 2^\Omega \to [0,1]$ as follows:

$$Lik_y(\theta) \stackrel{\text{def}}{=} \frac{lik_y(\theta)}{\sup_{\omega \in \Omega} lik_y(\omega)} = \frac{lik_y(\theta)}{lik_y(\hat\theta)} \quad (5)$$

$$Lik_y(A) \stackrel{\text{def}}{=} \sup_{\omega \in A} Lik_y(\omega) \text{ for } A \subseteq \Omega \quad (6)$$

where $\hat\theta$ is a maximum likelihood estimate of $\theta$.

The "extended" likelihood is not new. In fact, it was used by Neyman and Pearson in seminal papers published in 1928 [18] where their famous hypothesis test-ing theory was presented. The idea is to use the maximum likelihood estimate as the proxy for a set. Such a procedure is not only intuitively appealing, but it is also backed by various asymptotic optimality properties [13, 15, 19].

---

[1] There are two definitions of conditioning in possibility theory. One that is frequently found in possibility literature is called *ordinal* conditioning. The definition we use here is called *numerical* conditioning. See [7] for more details.

If in the process, one decides to focus on a particular subset of the parameter space $B \subseteq \Omega$, the effect of refocusing on the extended likelihood function could be express through what we call *conditioning*.

$$Lik_y(A|B) \stackrel{\text{def}}{=} \frac{Lik_y(A \cap B)}{Lik_y(B)}. \qquad (7)$$

We have a theorem whose proof consists of just verifying that the function $Lik_y$ satisfies the axioms of a possibility measure listed in eqs (1, 2, 3).

**Theorem 1** *The extended likelihood function $Lik_y$ is a possibility function.*

This technically obvious theorem is important because it establishes a direct link between possibility theory and the central concept in statistics. This relationship has been noticed for quite some time. The fact that the posterior possibility function calculated from a vacuous prior possibility and a random observation behaves as a likelihood function has been pointed out by Smets [21] if the updating process is required to satisfy certain postulates. Dubois et al. [6] show that possibility measures can be viewed as the supremum of a family of likelihood functions. Based on that they justify the min combination rule for possibility functions. We argue further that all relevant information about the unknown parameter that can be obtained from a partially specified probability model and an observation is a possibility measure. This view is essentially Shafer's idea [20] according to which statistical evidence is coded by belief functions whose focal elements are nested. This kind of belief function (*consonant*) in its *plausibility* form is a possibility function. Halpern and Fagin [12] argue for the representation of evidence by a *discrete probability function* that is obtained by normalizing likelihoods of the singletons so that they sum to unity. However, this probability, Halpern and Fagin concede, should be interpreted in neither frequentist nor subjectivist senses.

The identity of likelihood and possibility can be used in two directions. First, for possibility theory and its application in AI, the link with statistics boosts its relevance and validity. We argue that unlike the Bayesian approach, which assumes a completely specified distribution, the possibility function is the result of a partially specified model (true distribution is unknown but assumed to be in a $\mathcal{F}$) and an observed data. In other words, we have answered the question which is often raised in discussions: "Where does possibility information come from?" One answer is it comes from a partially specified probability model of a phenomenon and experimental observations. The derived possibility function encodes all relevant information that can be obtained from the experiment. Second, statistics can also benefit from recently developed techniques in an area that has a different motivation. Traditional problems in statistics may be re-examined in the light of new insights. In the rest of this paper, we study the statistical decision problem from the point of view of qualitative decision theory as described in [10, 11].

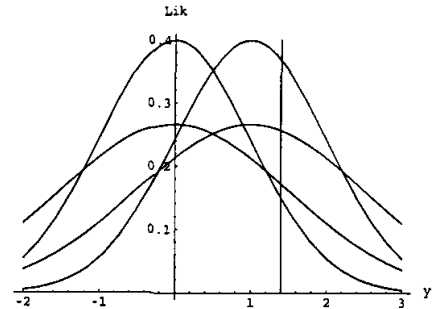

Figure 1: A Likelihood for $Y = 1.4$

**Example:** A random variable $Y$ is known to have a normal distribution. It is also known that mean $\mu \in \{0, 1\}$ and standard deviation $\sigma \in \{1, 1.5\}$. Suppose that value $y = 1.4$ is observed. Construct an extended likelihood function representing information about unknown parameters of distribution of $Y$.

Parameter $\theta = (\mu, \sigma)$. $\Omega = \{\omega_1, \omega_2, \omega_3, \omega_4\}$ with $\omega_1 = (0, 1)$, $\omega_2 = (0, 1.5)$, $\omega_3 = (1, 1)$, $\omega_4 = (1, 1.5)$.

|  | $\omega_1$ | $\omega_2$ | $\omega_3$ | $\omega_4$ |
|---|---|---|---|---|
| $lik_{1.4}(.)$ | 0.1497 | 0.1720 | 0.3683 | 0.2567 |
| $Lik_{1.4}(.)$ | 0.4066 | 0.4672 | 1.0 | 0.6970 |

| Events | $Lik_{1.4}(.)$ |
|---|---|
| $\mu = 0$ or $\{\omega_1, \omega_2\}$ | 0.4672 |
| $\mu = 1$ or $\{\omega_3, \omega_4\}$ | 1. |
| $\sigma = 1$ or $\{\omega_1, \omega_3\}$ | 1. |
| $\sigma = 1.5$ or $\{\omega_2, \omega_4\}$ | 0.6970 |
| $\mu = 0 | \sigma = 1$ | 0.4066 |
| $\mu = 1 | \sigma = 1$ | 1.0 |
| $\mu = 0 | \sigma = 1.5$ | 0.6703 |
| $\mu = 1 | \sigma = 1.5$ | 1.0 |

## 3 A Decision Theory with Likelihood Information Only

Let us review, in terms of likelihood semantics, our proposal for qualitative decision making with possibility theory [10, 11]. We assume a situation that includes a set $\Omega$ which can be interpreted as the set of hypotheses or the parameter space. Also assumed is a set of consequences or prizes $X$ that includes one best prize ($\overline{x}$) and one worst prize ($\underline{x}$). An action $d$ is a mapping $\Omega \to X$. In other words, the consequence of an action is determined by which hypothesis is true. The set of actions is denoted by $\mathcal{A}$. The uncertainty about hypotheses obtained from the observation is expressed by an extended likelihood function $\pi$. An action coupled with a uncertainty measure determines a *simple* lottery. Each lottery is a mechanism that delivers prizes with associated likelihoods. Formally, a lottery $L$ induced by $\pi$ and $d$ is a mapping from $X \to [0, 1]$ such that $L(x) = \pi(d^{-1}(x))$ for $x \in X$ where $d^{-1}$ is a set-valued inverse mapping of action $d$. For the remainder of this paper, we will denote a simple lottery by $[L(x_1)/x_1, L(x_2)/x_2, \ldots]$ with the convention that those $x_j$ for which $L(x_j) = 0$ are omitted. If we let prizes to be lotteries then we have *compound* lotteries. If a lottery involves only two prizes $\overline{x}$ and $\underline{x}$ (the best and the worst) as potential outcomes then it is called a *canonical* lottery. The set of canonical lotteries is denoted by $\mathcal{L}_c$.

We study a preference relation $\succeq$ on the set of lotteries $\mathcal{L}$. We postulate that $\succeq$ satisfies five axioms similar to those proposed by von Neumann and Morgenstern for the classical linear utility theory (in the form presented in [17]). They are as follows.

(A1) Order. $\succeq$ is reflexive, transitive, complete.

(A2) Reduction of compound lotteries.
Let $L = [\delta_1/L_1, \delta_2/L_2 \ldots \delta_k/L_k]$ and $L_i = [\kappa_{i1}/x_1, \kappa_{i2}/x_2, \ldots \kappa_{ir}/x_r]$ then $L \sim L_s$ where $L_s = [\kappa_1/x_1, \kappa_2/x_2, \ldots \kappa_r/x_r]$ with

$$\kappa_j = \max_{1 \leq i \leq k} \{\delta_i . \kappa_{ij}\} \quad (8)$$

(A3) Substitutability.
If $L_i \sim L'_i$ then $[\delta_1/L_1, \ldots \delta_i/L_i \ldots \delta_k/L_k] \sim [\delta_1/L_1, \ldots \delta_i/L'_i \ldots \delta_k/L_k]$

(A4) Continuity. For each $x_i \in X$ there is a $s \in \mathcal{L}_c$ such that $x_i \sim s$.

(A5) Qualitative monotonicity.

$$[\lambda/\overline{x}, \mu/\underline{x}] \succeq [\lambda'/\overline{x}, \mu'/\underline{x}] \text{ iff } (\lambda \geq \lambda') \& (\mu \leq \mu') \quad (9)$$

In [10, 11] we prove a representation theorem for the preference relation.

**Theorem 2** $\succeq$ *on $\mathcal{L}$ satisfies axioms A1 to A5 if and only if it is represented by a unique utility function $QU$ such that*

$$QU([\delta_1/L_1, \ldots, \delta_k/L_k]) = \max_{1 \leq i \leq k} \{\delta_i . QU(L_i)\} \quad (10)$$

where function $QU$ is a mapping $\mathcal{L} \to \mathcal{U}$ where $\mathcal{U} = \{\langle u, v \rangle | u, v \in [0, 1] \text{ and } \max(u, v) = 1\}$. The binary utility scale is formed by defining an order $\geq$ on $\mathcal{U}$ as follows

$$\langle \lambda, \mu \rangle \geq \langle \lambda', \mu' \rangle \text{ iff } \begin{cases} (1 \geq \lambda \geq \lambda' \ \& \ \mu = \mu' = 1) \vee \\ (\lambda = 1 \ \& \ \lambda' < 1) \vee \\ (\lambda = \lambda' = 1 \ \& \ \mu \leq \mu') \end{cases}$$

Multiplication and maximization are extended to pairs as follows ($\alpha, \beta, \gamma, \xi$ are scalar).

$$\alpha . \langle \beta, \gamma \rangle \stackrel{\text{def}}{=} \langle \alpha . \beta, \alpha . \gamma \rangle \quad (11)$$

$$\max\{\langle \alpha, \beta \rangle, \langle \gamma, \xi \rangle\} \stackrel{\text{def}}{=} \langle \zeta, \eta \rangle \quad (12)$$

where $\zeta = \max\{\alpha, \gamma\}$ and $\eta = \max\{\beta, \xi\}$

The expression in the right-hand side of eq.(10) is called the *qualitative expected* utility of the lottery in the left-hand side. As a result of theorem 2, we can compare the preference of lotteries by looking at their expected utility.

These five axioms were justified on intuitive grounds in [10, 11]. As we argued, the uncertainty information about the unknown parameter resulting from an experiment is captured by a possibility measure. Thus, the axioms can be justified indirectly in the context of statistical inference problem. However, we think it would be useful to reexamine them directly in terms of likelihood here.

Among the axioms, $A1$ and $A3$ are standard assumptions about a preference relation.

$A2$ is an implication of the conditioning operation. Suppose that the unknown parameter $\theta$ is a vector. We can think, for example, $\theta = (\gamma, \sigma)$. Let us consider a compound lottery $L = [\pi_1/L_1, \pi_2/L_2, \ldots, \pi_k/L_k]$ where $L_i = [\kappa_{i1}/x_1, \ldots, \kappa_{ir}/x_r]$ for $1 \leq i \leq k$. Underlying $L$, in fact, is a two-stage lottery. The first stage is associated with a scalar parameter $\gamma$. It accepts values $\gamma_1, \gamma_2, \ldots \gamma_k$ with likelihoods $\pi_1, \pi_2, \ldots \pi_k$ respectively. If $\gamma_i$ is the true value, the holder of $L$ is rewarded with simple lottery $L_i$ that, in turn, is associated with scalar parameter $\sigma$ that accepts $\sigma_{o_i(1)}, \sigma_{o_i(2)}, \ldots \sigma_{o_i(r)}$ with likelihoods $\kappa_{i1}, \kappa_{i2}, \ldots \kappa_{ir}$ where $o_i$ is a permutation of $(1, 2, 3, \ldots r)$. When $\sigma_{o_i(j)}$ obtains, the holder is rewarded with prize $x_j$. We can view this lottery from a different angle. That is $L$ is a lottery that delivers $x_j$ in case tuple $<\gamma_i \sigma_{o_i(j)}>$ is the true value of $\theta$. Thus, the extended likelihood $<\gamma_i \sigma_{o_i(j)}>$ is $\pi_i . \kappa_{ij}$ since $\kappa_{ij}$ is the conditional likelihood on $\gamma_i$. The set of tuples for which $x_j$ is delivered is $\{<\gamma_i \sigma_{o_i(j)}> | 1 \leq i \leq k\}$.

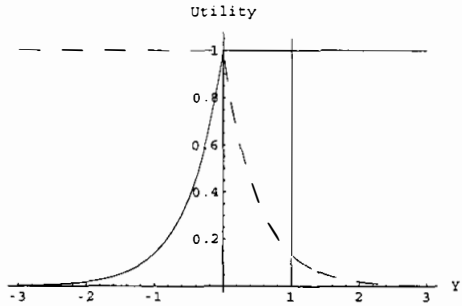

Figure 2: Binary utility from observation

Thus, the extended likelihood associated with prize $x_j$ in lottery $L$ is $\max\{\pi_i.\kappa_{ij}|1 \leq i \leq k\}$.

The continuity axiom ($A4$) requires that for any consequence $x \in X$ there is a canonical lottery $c = (\lambda_1/\overline{x}, \lambda_2/\underline{x})$ such that $x \sim c$. For clarity, let us assume that $\overline{x} = 1$, $\underline{x} = 0$. For any $x \in [0,1]$, we need to find a canonical lottery $c$ equivalent to $x$. We propose a *likelihood gamble* for that purpose. By doing so, we provide an operational semantic for the concept of binary utility.

Suppose that random variable $Y$ is generated from one of two normal distributions[2] that have the same standard deviation $\sigma = 1$ but with two different means, either $-1$ or $1$, i.e., $Y \sim N(-1,1)$ or $Y \sim N(1,1)$. Next you are informed that a randomly observed value of $Y$ is $y$. What is the most you would be willing to pay for a gamble that pays \$1 if the true mean $\mu$ is $-1$ and nothing if the true mean is 1? If the answer is $x$, then it can be shown that for you

$$x \sim [Lik_y(-1)/1, Lik_y(1)/0] \tag{13}$$

Figure 2 plots functions $\lambda_1(y) = Lik_y(-1)$ (the dashed line) and $\lambda_2(y) = Lik_y(1)$.

This betting procedure is quite similar to the test used in practice to identify the unary utility of a monetary value. But there are important differences. First, instead of using a random variable with a known distribution (e.g., tossing a fair coin or rolling a dice), we use for our test a partially specified model together with a randomly generated observation. We let the decision maker bet on a gamble tied to a situation that by nature, is a hypothesis testing problem. The rewards in this gamble are tied not to an observed realization of the random variable as in an ordinary gamble, but instead, the rewards are conditioned on the true value of the parameter underlying the random variable.

---

[2]The use of normal distributions is for illustration purpose only. Any other distribution will do.

We just showed that binary utility elicitation is done by betting on a likelihood gamble. We assume that the betting behavior of the decision maker is consistent and rational in some way. For example, paying \$0.20 for the gamble if $Y = -3$ and paying \$0.70 if $Y = 1$ is considered irrational. The constraint we impose is formulated as axiom $A5$, qualitative monotonicity ($\overline{x} = 1$ and $\underline{x} = 0$). In other words, we require that the price for the lottery $[\lambda/1, \mu/0]$ is no less than the price for $[\lambda'/1, \mu'/0]$ if the condition on the right hand of eq.(9) is satisfied.

Since the likelihood gamble, as noted, is a hypothesis testing problem. We show that axiom $A5$ is supported by the traditional statistical inference methods namely classical, Bayesian and likelihood. Let us denote by $y$ and $y'$ the values of $Y$ that associate with gambles $[\lambda/1, \mu/0]$ and $[\lambda'/1, \mu'/0]$ respectively. Referring to the graphs in figure 2, it is easy to check the fact that the condition on the right hand of eq.(9) is equivalent to $y \leq y'$. We want to show that the confidence in favor of $\mu = -1$ is higher for $y$ than for $y'$ no matter which methods are used.

For the significance test method applied for testing $\mu = 1$ against $\mu = -1$, the $p$-value for rejecting $\mu = 1$ in favor of $\mu = -1$ provided by data $y$ is $P_{\mu=1}(Y \leq y)$ that is smaller than the $p$-value corresponding to data $y'$ which is $P_{\mu=1}(Y \leq y')$.

The likelihood reasoning [8] compares the hypotheses by looking at their likelihood ratios. It is obvious that

$$\frac{lik_y(-1)}{lik_y(1)} \geq \frac{lik_{y'}(-1)}{lik_{y'}(1)} \tag{14}$$

If the decision maker is Bayesian and assumes the prior probability $P(\mu = -1) = p$, then (s)he will calculate posterior probabilities

$$P(\mu = -1|y) = \frac{p.Lik_y(-1)}{p.Lik_y(-1) + (1-p).Lik_y(1)} \tag{15}$$

From $Lik_y(-1) \geq Lik_{y'}(-1)$ and $Lik_y(1) \leq Lik_{y'}(1)$, we have $P(\mu = -1|y) \geq P(\mu = -1|y')$. Let us denote the expected payoffs of the likelihood gambles in two cases $Y = y$ and $Y = y'$ by $V_y$ and $V_{y'}$. $V_y = P(\mu = -1|y).1 + P(\mu = 1|y).0 = P(\mu = -1|y) \geq P(\mu = -1|y') = V_{y'}$.

We want to invoke one more, in our opinion the most important, justification for $A5$. This justification is based on the concept of the first order stochastic dominance (FSD).

Without being a strict Bayesian, we do not assume to know the prior probability of $P(\mu = -1)$. We can model this situation by viewing the prior probability of $\mu = -1$ as a random variable $\rho$ taking value in the unit interval. However, we do not assume to know the distribution of $\rho$. We will write the expected payoff of the likelihood gamble by $V_y(\rho)$. Since it is a function of $\rho$, $V_y(\rho)$ is a random variable. We want to compare two r.v. $V_y(\rho)$ and $V_{y'}(\rho)$.

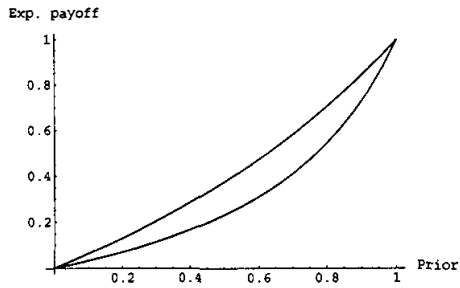

Figure 3: Expected payoffs for two observations

One of the methods used for comparing random variables is based on the concept of *stochastic dominance* (SD). We are interested in the first degree stochastic dominance (FSD). This concept has been used extensively in economics, finance, statistics and other areas [16]. Suppose $X$ and $Y$ are two distinct random variables with the cumulative distributions $F$ and $G$ respectively. We say that $X$ stochastically dominates (first degree) $Y$ (write $XD_1Y$) iff $F(x) \leq G(x)$ $\forall x$. Since $X$ and $Y$ are distinct, strict inequality must hold for at least one value $x$. FSD is important because of the following equivalence $XD_1Y$ iff $E(u(X)) \geq E(u(Y))$ $\forall u \in U$ where $U$ the class of non-decreasing utility functions. $E(u(X))$ is the expected utility of $X$ under utility function $u$.

Let us consider the relation between $V_y(\rho)$ and $V_{y'}(\rho)$. We have the following lemma that relates the expected payoffs and the likelihood pairs. The proof is omitted.

**Lemma 1** *For a given value $\rho \notin \{0,1\}$, $V_y(\rho) > V_{y'}(\rho)$ iff*

$$\langle Lik_y(-1), Lik_y(1)\rangle > \langle Lik_{y'}(-1), Lik_{y'}(1)\rangle \quad (16)$$

In figure 3, the lower curve is the graph for $V_{0.60}(\rho)$ (at $Y = .6$ the extended likelihood pair is $\langle .3011, 1.\rangle$) and the upper curve is the graph for $V_{0.26}(\rho)$ ($\langle .5945, 1.\rangle$).

Now, for any $v \in (0,1)$, let us denote the roots of equations $V_y(\rho) = v$ and $V_{y'}(\rho) = v$ by $\rho_v$ and $\rho'_v$ respectively i.e., $V_y(\rho_v) = v$ and $V_{y'}(\rho'_v) = v$. If $\langle \lambda, \mu \rangle > \langle \lambda', \mu' \rangle$ then by lemma 1 $V_y(\rho_v) > V_{y'}(\rho_v)$. Therefore, $V_{y'}(\rho'_v) > V_{y'}(\rho_v)$. Because $V_{y'}(\rho)$ is strictly increasing, we infer $\rho_v < \rho'_v$.

Since $V_y(\rho)$ and $V_{y'}(\rho)$ are increasing, $P(V_y(\rho) \leq v) = P(\rho \leq \rho_v)$ and $P(V_{y'}(\rho) \leq v) = P(\rho \leq \rho'_v)$. Because $\rho_v < \rho'_v$, $P(V_y(\rho) \leq v) \leq P(V_{y'}(\rho) \leq v)$. This last inequality means $V_y(\rho)$ $D_1$ $V_{y'}(\rho)$. Thus, we have the following theorem

**Theorem 3** *The order on binary utility is the order of first degree stochastic dominance.*

In summary, we show that there are compelling arguments justifying the five proposed axioms.

## 4 A Decision-Theoretic Approach to Statistical Inference with Likelihood

We will review the decision theoretic approach to statistical inference. We assume given the set of alternative actions denoted by $\mathcal{A}$, the sample space of $Y$ by $\mathcal{Y}$. A *loss* $L(a, \theta)$ measures the loss that arises if we take action $a$ and the true value of the parameter is $\theta$. A decision rule is a mapping $\delta : \mathcal{Y} \to \mathcal{A}$, that is for an observation $y$ the rule recommends an action $\delta(y)$. The *risk function* of a decision rule $\delta$ at parameter value $\theta$ is defined as

$$R(\delta(Y), \theta) \stackrel{\text{def}}{=} E_\theta L(\delta(Y), \theta) = \int_{\mathcal{Y}} L(\delta(y), \theta) p_\theta(y) \quad (17)$$

The risk function measures the average loss by adopting the rule $\delta$ in case $\theta$ is the true value.

The further use of risk functions depends on how much information we assume is available. For each point in the parameter space, there is a value of risk function. In case no a priori information about parameter exists, Wald [22] advocated the use of minimax rule which minimizes the worst risk that could be attained by a rule.

$$\delta^* = \arg \min_{\delta \in \Delta} \max_{\theta \in \Omega} R(\delta, \theta) \quad (18)$$

where $\Delta$ is the set of decision rules. $\delta^*$ is called the *minimax solution*.

If we assume, as the Bayesian school does, the existence of a prior distribution of the parameter then the risk could be averaged out to one number called *Bayes risk*

$$r(\delta) = E_\rho R(\delta, \theta) \quad (19)$$

where $\rho$ is prior distribution of $\theta$. Then the optimal rule is one that minimizes the Bayes risk which is called the *Bayes solution*.

Wald [22] pointed out the relationship between the two rules. There exists a prior distribution $\rho$ called "the least favorable" for which Bayes solution is the minimax solution.

In our view, both solution concepts are not entirely satisfactory. Bayes solution is made possible by assuming the existence of prior distribution, which is often a leap of faith. Although in highly idealized situation, prior probability may be extracted from preference or betting behavior, in practice, there are many factors that violate conditions necessary for such deduction.

We have two problems with Wald's proposal. First, the risk function has no *special* role for the actually observed data. Thus, Wald's proposal does not respect the principle of conditionality. That violation leads to the second problem of minimax rule. The rule ignores information about the parameter that is obtained from the data. One can argue that the minimax rule is being cautious. But it is, in our opinion, too cautious. There is information about the parameter $\theta$ provided by the observed data, namely, likelihood information, that must be utilized in making a decision.

We propose the following solution. For a given $y \in \mathcal{Y}$, an extended likelihood function $Lik_y$ is calculated. That allows comparison of actions on the basis of their qualitative expected utilities, or more exactly, utilities of lotteries induced by the actions and the likelihood function. For an action $d$ and an observation $y$, denote $L_d(y)$ the lottery generated. The selected action given observation $y$ is

$$d^*(y) = \arg\sup_{d \in \mathcal{A}} QU(L_d(y)) \qquad (20)$$

where $sup^3$ is the operation taking maximum element according to the binary order. Equation (20) also defines a decision rule in the sense that for each point in the sample space, it gives an action. We call such a decision rule the *likelihood solution* for obvious reasons.

## 5  An Illustrative Example

The following example is adapted from [1]. The manufacturer of small travel clocks which are sold through a chain of department stores agrees to service any clock once only if it fails to operate satisfactorily in the first year of purchase. For any clock, a decision must be made on whether to merely clean it or replace the works, i.e., the set of actions $\mathcal{A} = \{a_1, a_2\}$ where $a_1$ denotes cleaning the clock, and $a_2$ denotes immediately replacing the works.

Let us assume that there are only two possible faults i.e., $\Omega = \{\theta_1, \theta_2\}$ where $\theta_1$ means there is the need for cleaning and $\theta_2$ means the clock has been physically damaged and the works need replacement. Utility and loss functions are given in the following table. The relationship between utility and loss is through equation Loss = 1 − Utility.

---

[3]In contrast to *max* defined in eq. 12 that operates on scalars.

| $u(a,\theta)$ | $\theta_1$ | $\theta_2$ | $L(a,\theta)$ | $\theta_1$ | $\theta_2$ |
|---|---|---|---|---|---|
| $a_1$ | .8 | .3 | $a_1$ | .2 | .7 |
| $a_2$ | .5 | .5 | $a_2$ | .5 | .5 |

The loss table is roughly estimated from the fact that cleaning a clock costs $0.20 and replacing the works costs $0.50. If the policy is to replace the works for every clock needing service then the cost is $0.50 no matter which problem is present. If the policy is to clean a clock first, if the state is $\theta_1$ then the service costs $0.20, however in the case of physical damage then cleaning alone obviously does not fix the problem and the manufacturer ends up replacing the works also. Thus the total cost is $0.70.

The manufacturer can ask the customer to provide a symptom of malfunction when a clock is sent to the service center. The symptom can be viewed as observation. Assume the sample space $\mathcal{Y} = \{y_1, y_2, y_3\}$ where $y_1$ means "the clock has stopped operating", $y_2$ - "the clock is erratic in timekeeping and $y_3$ - "clock can only run for a limited period". Such information gives some indication about $\theta$ that is expressed in terms of likelihood

| $Lik_y(\theta)$ | $y_1$ | $y_2$ | $y_3$ |
|---|---|---|---|
| $\theta_1$ | .1 | .4 | .5 |
| $\theta_2$ | .7 | .2 | .1 |

For each point in the sample space, you can either choose $a_1$ or $a_2$, so there are 8 possible decision rules in total. Each decision rule specifies an action given an observation.

| | $\delta_1$ | $\delta_2$ | $\delta_3$ | $\delta_4$ | $\delta_5$ | $\delta_6$ | $\delta_7$ | $\delta_8$ |
|---|---|---|---|---|---|---|---|---|
| $y_1$ | $a_1$ | $a_1$ | $a_1$ | $a_1$ | $a_2$ | $a_2$ | $a_2$ | $a_2$ |
| $y_2$ | $a_1$ | $a_1$ | $a_2$ | $a_2$ | $a_1$ | $a_1$ | $a_2$ | $a_2$ |
| $y_3$ | $a_1$ | $a_2$ | $a_1$ | $a_2$ | $a_1$ | $a_2$ | $a_1$ | $a_2$ |

We calculate the risk function values for each rule and parameter value in the following table

| $R_{ij}$ | $\delta_1$ | $\delta_2$ | $\delta_3$ | $\delta_4$ | $\delta_5$ | $\delta_6$ | $\delta_7$ | $\delta_8$ |
|---|---|---|---|---|---|---|---|---|
| $\theta_1$ | .2 | .35 | .32 | .47 | .23 | .38 | .35 | .50 |
| $\theta_2$ | .7 | .68 | .66 | .64 | .56 | .54 | .52 | .50 |

Notice that there is no rule which is superior to all other for both values of $\theta$. Wald's minimax solution is $\delta_8$.

If we assume prior distribution of $\theta$ then we can calculate the Bayes risks for the rules. For example if prior probability $p(\theta_1) = .7$, then Bayes risks $r_{.7}(\delta_i)$ are

| $\delta_1$ | $\delta_2$ | $\delta_3$ | $\delta_4$ | $\delta_5$ | $\delta_6$ | $\delta_7$ | $\delta_8$ |
|---|---|---|---|---|---|---|---|
| .35 | .449 | .442 | .541 | .329 | .428 | .401 | .50 |

In this case, the Bayes solution is $\delta_5$. It is easy to understand why the Bayes solution will be different if the prior $p(\theta_1)$ changes. Indeed, a sensitivity analysis shows that Bayes solution is

| Bayes solution | When |
|---|---|
| $\delta_1$ | $p(\theta_1) \geq .824$ |
| $\delta_5$ | $.250 \leq p(\theta_1) \leq .824$ |
| $\delta_7$ | $.118 \leq p(\theta_1) \leq .250$ |
| $\delta_8$ | $p(\theta_1) \leq .118$ |

In our approach, suppose the betting procedure produces the following preference equivalence between monetary values and binary utility.

| Unary utility | Binary utility |
|---|---|
| .8 | $\langle 1, .25 \rangle$ |
| .5 | $\langle 1, 1 \rangle$ |
| .3 | $\langle .43, 1 \rangle$ |

Given observation $y_1$, the extended likelihood function is $Lik_{y_1}(\theta_1) = .14$ and $Lik_{y_1}(\theta_2) = 1$. Action $a_1$ is associated with lottery $L_{a_1}(y_1) = [.14/\langle 1, .25\rangle, 1/\langle .43, 1\rangle]$ whose qualitative expected utility is calculated as

$$\begin{aligned} QU(L_{a_1}(y_1)) &= \max\{.14\langle 1, .25\rangle, 1\langle .43, 1\rangle\} \\ &= \max\{\langle .14, .035\rangle, \langle .43, 1\rangle\} \\ &= \langle .43, 1\rangle \end{aligned}$$

Action $a_2$ is associated with lottery $L_{a_2}(y_1) = [.14/\langle 1,1\rangle, 1/\langle 1,1\rangle]$ whose qualitative expected utility $QU(L_{a_2}(y_1)) = \langle 1, 1\rangle$. Thus, given $y_1$, we have $a_2 \succ_{y_1} a_1$ i.e., $a_2$ is strictly prefered to $a_1$.

Given observation $y_2$, the extended likelihood function is $Lik_{y_2}(\theta_1) = 1$ and $Lik_{y_2}(\theta_2) = .5$. We calculate the qualitative expected utility for $a_1$ is $QU(L_{a_1}(y_2)) = \langle 1, .5\rangle$ and for $a_2$ $QU(L_{a_2}(y_2)) = \langle 1, 1\rangle$. Thus, $a_1 \succ_{y_2} a_2$.

Given observation $y_3$, the extended likelihood function is $Lik_{y_2}(\theta_1) = 1$ and $Lik_{y_2}(\theta_2) = .2$. The qualitatively expected utility for $a_1$ is $QU(L_{a_1}(y_3)) = \langle 1, .25\rangle$ and for $a_2$ remains $QU(L_{a_2}(y_3)) = \langle 1, 1\rangle$. Thus, $a_1 \succ_{y_3} a_2$.

In summary, our approach suggests $\delta_5$ as the likelihood solution.

Let us make an informal comparison of likelihood solution with minimax and Bayes solutions. In this example, likelihood solution $\delta_5$ is different from the minimax solution $\delta_8$. It is because, as we noted, minimax solution ignores the uncertainty generated by an observation while likelihood solution does not. In that sense, likelihood solution is more information efficient.

If the prior probability $p(\theta_1) = .7$, then the Bayes solution is $\delta_5$ the same as the likelihood solution. If prior probability is available, one can argue that Bayes solution is *the* optimal one. However, the "optimality" of Bayes solution comes at a cost. The decision maker must possess the prior probability. This condition can be satisfied either at some monetary cost (doing research, or buying from those who know) or the decision maker can just assume some prior distribution. In the latter case, the cost is a compromise of solution optimality. One can extend the concept of Bayes solution to include the sensitivity analysis. This certainly helps decision maker by providing a frame of reference. But sensitivity analysis itself does not constitute any basis for knowing the prior probability.

The fact that likelihood solution $\delta_5$ coincides with Bayes solution that corresponds to a largest interval of prior should not be exaggerated. Different unary-binary utility conversions may lead to different likelihood solution. However, it is important to note that there is a fundamental reason for the agreement. As we have pointed out, the axioms $A1$ to $A5$ on which likelihood solution is based, are structurally similar to those in [17]. In that work, Luce and Raiffa used those axioms to justify linear utility concept which ultimately is a basis for Bayes solution. Thus, at a foundational level, optimality of likelihood solution could be justified in the same way as the optimality for Bayes solution although the two optimality concepts are obviously different. We argue that the question of which optimality has precedence over the other depends on how much information is available.

## 6 Summary and Conclusion

In this paper, we take the fundamental position that all relevant information about the unknown parameter is captured by the likelihood function. This position is justified on the basis of the Likelihood Principle and various asymptotic optimality properties of the maximum likelihood procedure. We show that such a likelihood function is basically a possibility measure.

We re-examine the axioms of our decision theory with possibility calculus in terms of the likelihood semantics. We provide a betting procedure that determines, for a decision maker, the binary utility of a monetary value. This test justifies the continuity axiom. We also justify the monotonicity axiom $A5$ in terms of first order stochastic dominance.

We propose a new *likelihood* solution to a statistical decision problem. The solution is based on maximizing the qualitative expected utility of an action given the likelihood function associated with a point in the sample space. The likelihood solution is sandwiched by Wald's minimax solution on one side and the Bayes solution on the other side. Compared to the minimax solution, the likelihood solution is more information efficient. Compared to the Bayesian solution, the likelihood solution does not require a prior probability dis-tribution of the unknown parameter.

## Acknowledgment

We thank anonymous referees for comments, reference and suggestions to improve this paper.